\pdfoutput=1
\documentclass[11pt,a4paper]{article}
\usepackage[hyperref]{acl2019}
\hypersetup{
    pdfsubject={Fine-Grained Entity Typing in Hyperbolic Space},
    pdftitle={Fine-Grained Entity Typing in Hyperbolic Space},
    pdfauthor={Federico Lopez},
    pdfkeywords={natural language processing, fine-grained entity typing, entity typing, hyperbolic space, hierarchy}
}
\usepackage{times}
\usepackage{latexsym}

\usepackage{url}
\usepackage{amsmath}
\usepackage{amssymb}

\usepackage{caption}
\usepackage{tabularx}
\usepackage{xcolor}
\usepackage{svg}

\usepackage{graphicx}
\graphicspath{{./fig/}{./fig_raw/}}
\usepackage{booktabs}
\usepackage{adjustbox}
\usepackage{multirow}

\usepackage{subfig}

\usepackage{floatrow}
\newfloatcommand{capbtabbox}{table}[][\FBwidth]

\aclfinalcopy 

\title{Fine-Grained Entity Typing in Hyperbolic Space}

\author{Federico L\'opez* \\
  \\\And
  Benjamin Heinzerling \\
  *Research Training Group AIPHES \\
  Heidelberg Institute for Theoretical Studies \\
  {\tt firstname.lastname@h-its.org} \\
  \\\And
  Michael Strube\\
 }
 
\date{}

\begin{document}
\maketitle
\begin{abstract}

How can we represent hierarchical information present in large type inventories for entity typing? We study the ability of hyperbolic embeddings to capture hierarchical relations between mentions in context and their target types in a shared vector space. We evaluate on two datasets and investigate two different techniques for creating a large hierarchical entity type inventory: from an expert-generated ontology and by automatically mining type co-occurrences. We find that the hyperbolic model yields improvements over its Euclidean counterpart in some, but not all cases. Our analysis suggests that the adequacy of this geometry depends on the \textit{granularity} of the type inventory and the way hierarchical relations are inferred.\footnote{Code available at: \url{https://github.com/nlpAThits/figet-hyperbolic-space}}



\end{abstract}

\section{Introduction}
\label{sec:intro}

Entity typing classifies textual mentions of entities according to their semantic class. The task has progressed from finding company names \cite{rau1991extracting}, to recognizing coarse classes (\textit{person}, \textit{location}, \textit{organization}, and \textit{other}, \citealp{sang2003conll}), to fine-grained inventories of about one hundred types, with finer-grained types proving beneficial in applications such as relation extraction \cite{yaghoobzadeh2017relation} and question answering \cite{YavuzD16-1015}. The trend towards larger inventories has culminated in \textit{ultra-fine} and \textit{open} entity typing with thousands of classes \cite{choi2018ultra,zhou2018zeroshotET}.

\begin{figure}[!t]
\begin{floatrow}
\capbtabbox{
\centering
\scalebox{0.7}{
\begin{tabular}{ll}
\specialrule{.1em}{.05em}{.05em}
\textbf{Sentence} & \textbf{Annotation} \\
\hline
\begin{tabular}[c]{@{}l@{}}...when the \\ \textbf{president} said...\end{tabular} & \begin{tabular}[c]{@{}l@{}}politician, \\ president\end{tabular} \\
\hline
\begin{tabular}[c]{@{}l@{}}...during the \\ negotiation, \textbf{he}...\end{tabular} & \begin{tabular}[c]{@{}l@{}}politician, \\ diplomat\end{tabular} \\
\hline
\begin{tabular}[c]{@{}l@{}}...after the last \\ meeting, \textbf{she}...\end{tabular} & \begin{tabular}[c]{@{}l@{}}politician, \\ president\end{tabular} \\
\hline
\begin{tabular}[c]{@{}l@{}}...the \textbf{president} \\ argued...\end{tabular} & \begin{tabular}[c]{@{}l@{}}politician, \\ president\end{tabular} \\
\specialrule{.1em}{.05em}{.05em}
\end{tabular}
}
}


\subfloat{\label{fig:mdright}{\includegraphics[height=3cm,keepaspectratio]{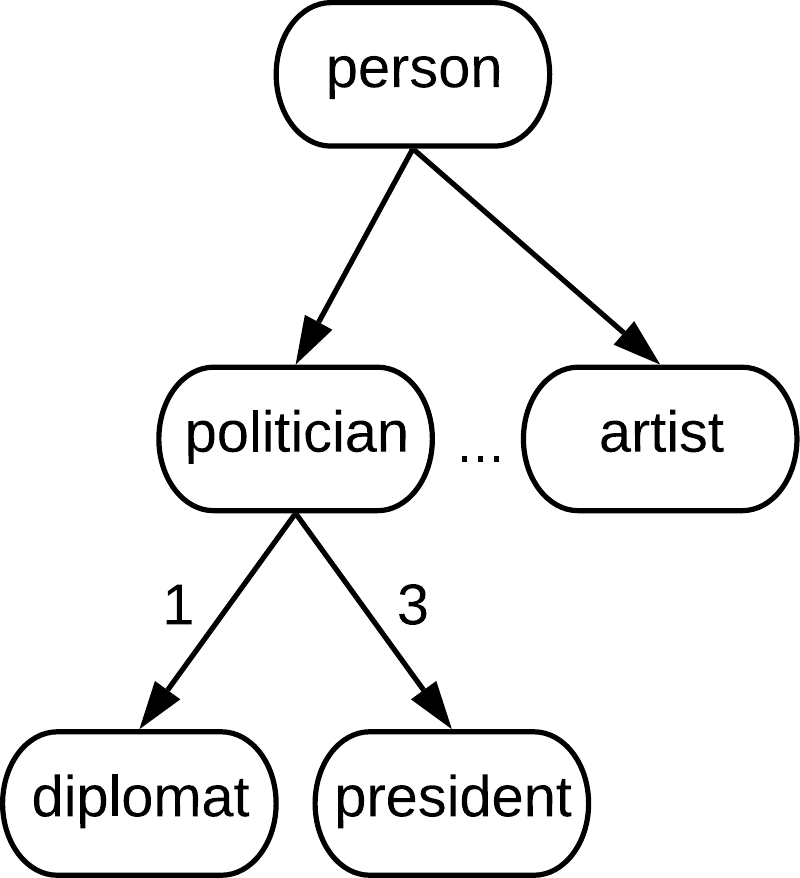}}}
\caption{Examples of annotations and hierarchical type inventory with co-occurrence frequencies.}
\label{fig:freq-hier}
\end{floatrow}
\end{figure}


However, large type inventories pose a challenge for the common approach of casting entity typing as a multi-label classification task \cite{yogatama2015embeddings, shimaoka2016attentive}, since exploiting inter-type correlations becomes more difficult as the number of types increases.
A natural solution for dealing with a large number of types is to organize them in hierarchy ranging from general, \textit{coarse} types such as ``person'' near the top, to more specific, \textit{fine} types such as ``politician'' in the middle, to even more specific, \textit{ultra-fine} entity types such as ``diplomat'' at the bottom (see Figure~\ref{fig:freq-hier}).
By virtue of such a hierarchy, a model learning about diplomats will be able to transfer this knowledge to related entities such as politicians.


Prior work integrated hierarchical entity type information by formulating a hierarchy-aware loss \cite{ren2016LabelNoiseReduction, murty2018hierarchicaLosses, xuBarbosa2018hierarchyAware}
or by representing words and types in a joint Euclidean embedding space \cite{shimaoka2017neural, abhishek2017jointLearning}. 
Noting that it is impossible to embed arbitrary hierarchies in Euclidean space, \newcite{nickel2017poincare} propose hyperbolic space as an alternative and show that hyperbolic embeddings accurately encode hierarchical information.
Intuitively (and as explained in more detail in Section~\ref{sec:background}), this is because distances in hyperbolic space grow exponentially as one moves away from the origin, just like the number of elements in a hierarchy grows exponentially with its depth.

While the intrinsic advantages of hyperbolic embeddings are well-established, their usefulness in downstream tasks is, so far, less clear.
We believe this is due to two difficulties: First, incorporating hyperbolic embeddings into a neural model is non-trivial since training involves optimization in hyperbolic space. Second, it is often not clear what the best hierarchy for the task at hand is. 

In this work, we address both of these issues. Using ultra-fine grained entity typing \cite{choi2018ultra} as a test bed, we first show how to incorporate hyperbolic embeddings into a neural model (Section~\ref{sec:ETinHyperbolicSpace}). 
Then, we examine the impact of the hierarchy, comparing hyperbolic embeddings of an expert-generated ontology to those of a large, automatically-generated one (Section~\ref{sec:hierarchies}).
As our experiments on two different datasets show (Section~\ref{sec:experiments}), hyperbolic embeddings improve entity typing in some but not all cases, suggesting that their usefulness depends both on the type inventory and its hierarchy.
In summary, we make the following contributions:
\begin{enumerate}
	\item We develop a fine-grained entity typing model that embeds both entity types and entity mentions in hyperbolic space.
	\item We compare two different entity type hierarchies, one created by experts (WordNet) and one  generated automatically, and find that their adequacy depends on the dataset.
	\item We study the impact of replacing the Euclidean geometry with its hyperbolic counterpart in an entity typing model, finding that the improvements of the hyperbolic model are noticeable on \textit{ultra-fine} types.
\end{enumerate}

\section{Background: Poincar\'e Embeddings}
\label{sec:background}


\begin{figure}[!t]
\centering
\subfloat[Euclidean Space.]{\label{fig:sub1}{\includegraphics[width=.45\linewidth,keepaspectratio]{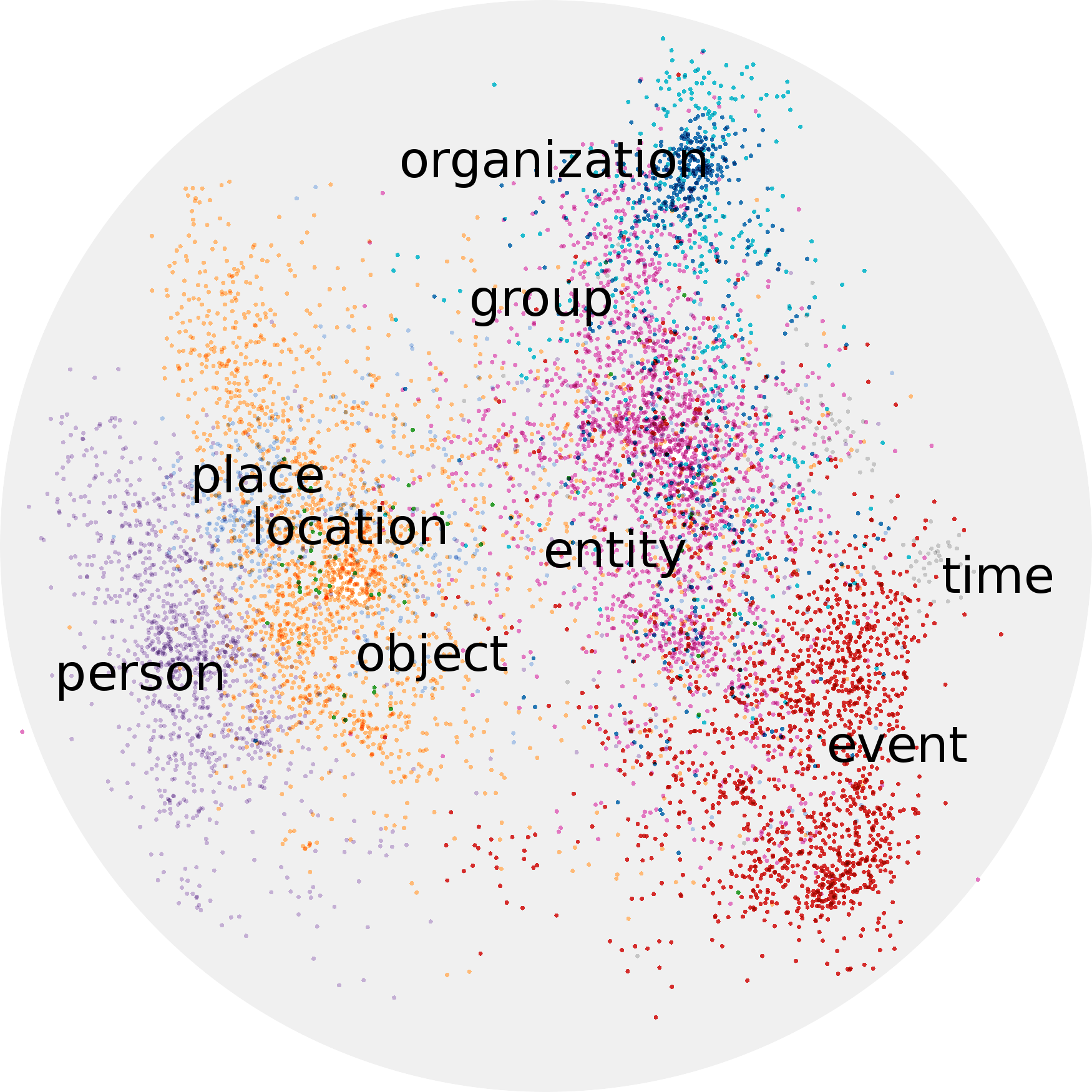}}}\hfill
\subfloat[Hyperbolic Space.]{\label{fig:sub2}{\includegraphics[width=.45\linewidth,keepaspectratio]{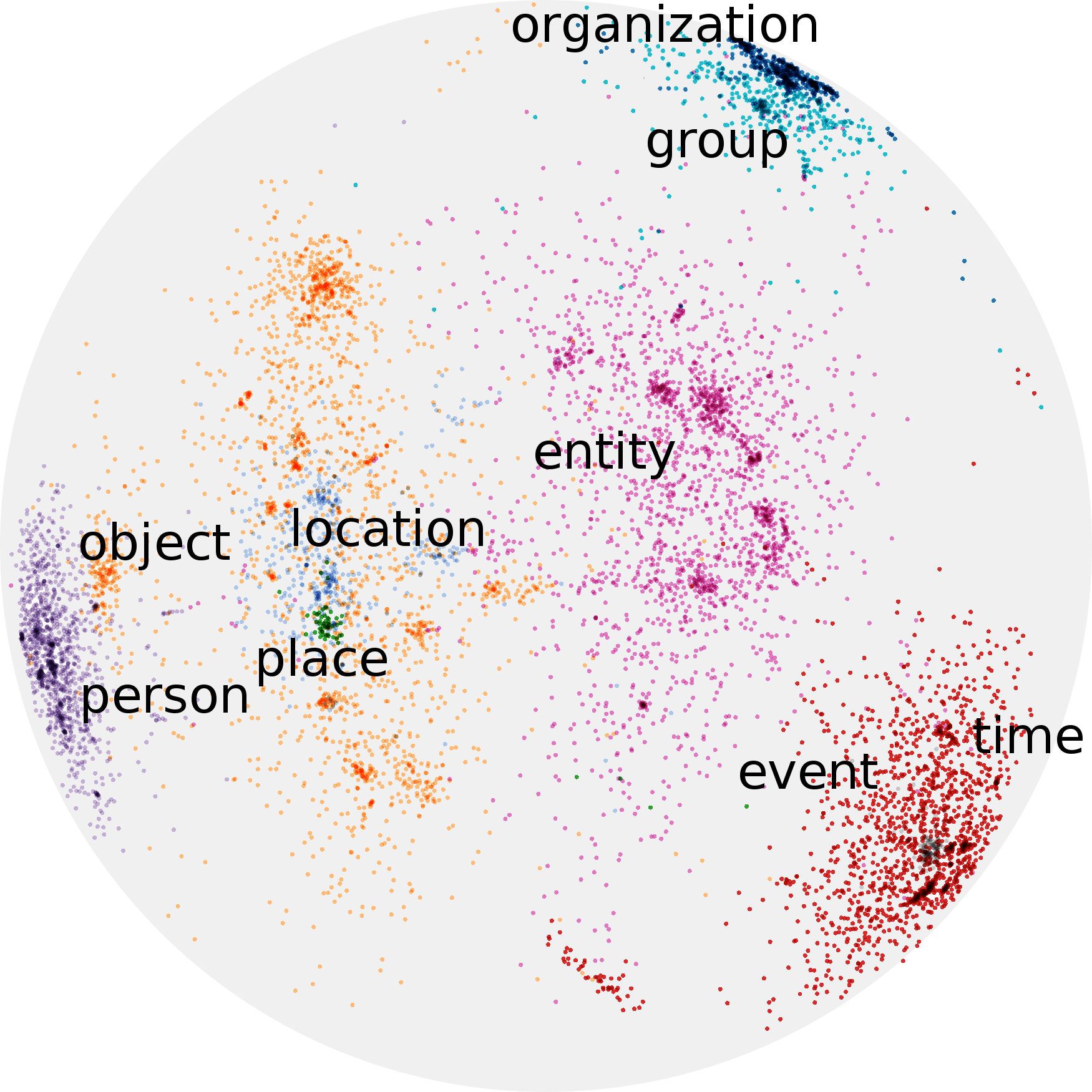}}}
\caption{Type inventory of the Ultra-Fine dataset aligned to the WordNet noun hierarchy and projected on two dimensions in different spaces.}
\label{fig:wnet}
\end{figure}


\begin{figure*}[t]
\centering
	\includegraphics[trim={0 0.8cm 0 0}, height=5cm, keepaspectratio]{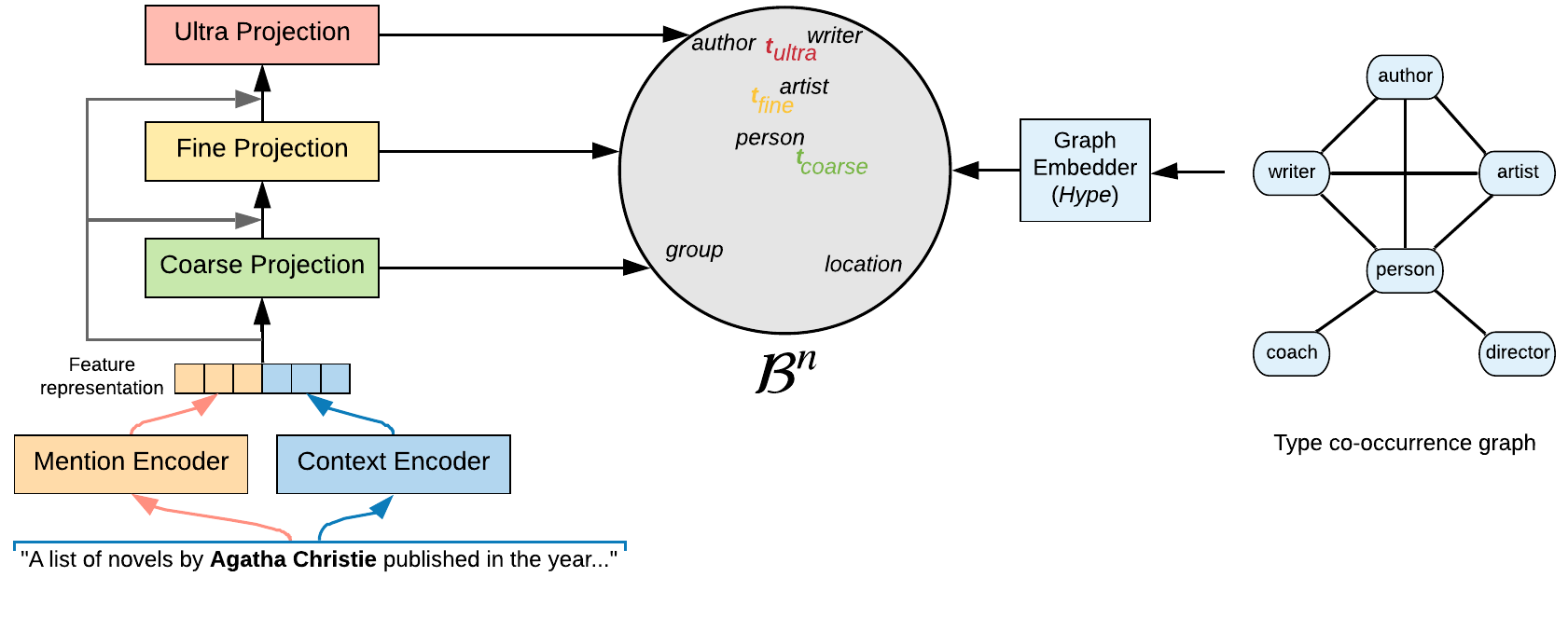}
	\subfloat[\label{fig:model-proj} Projection layers.]{\hspace{.5\linewidth}}
	\subfloat[\label{fig:model-hier} Incorporation of hierarchical information.]{\hspace{.5\linewidth}}
	\caption{Overview of the proposed model to predict types of a mention within its context.}
	\label{fig:overview}
\end{figure*}


Hyperbolic geometry studies non-Euclidean spaces with constant negative curvature. Two-dimensional hyperbolic space can be modelled as the open unit disk, the so-called \emph{Poincar\'e disk}, in which the unit circle represents infinity, i.e.,\ as a point approaches infinity in hyperbolic space, its norm approaches one in the Poincar\'e disk model. In the general $n$-dimensional case, the disk model becomes the Poincar\'e ball \cite{Chamberlain2017} $\mathcal{B}^n = \{x \in \mathbb{R}^n \mid \|x\| < 1\}$, where $\| \cdot \|$ denotes the Euclidean norm. In the Poincar\'e model the distance between two points $u,v \in \mathcal{B}^n$ is given by:%
\begin{equation}
\label{eq:hyperDist}
\small
d_{H}(\textbf{u}, \textbf{v}) = \operatorname{arcosh}(1 + 2 \frac{\|\textbf{u} - \textbf{v}\|^2}{(1 - \|\textbf{u}\|^2)(1 - \|\textbf{v}\|^2)})
\end{equation}%
If we consider the origin $O$ and two points, $x$ and $y$, moving towards the outside of the disk, i.e.  $\|x\|, \|y\| \to 1$, the distance $d_{H}(x, y)$ tends to $d_{H}(x, O) + d_{H}(O, y)$. That is, the path between $x$ and $y$ is converges to a path through the origin. This behaviour can be seen as the continuous analogue to a (discrete) tree-like hierarchical structure, where the shortest path between two sibling nodes goes through their common ancestor.

As an alternative intuition, note that the hyperbolic distance between points grows exponentially as points move away from the center. This mirrors the exponential growth of the number of nodes in trees with increasing depths, thus making hyperbolic space a natural fit for representing trees and hence hierarchies \cite{krioukov2010hypernetworks,nickel2017poincare}.

By embedding hierarchies in the Poincar\'e ball so that items near the top of the hierarchy are placed near the origin and lower items near infinity (intuitively, embedding the ``vertical'' structure), and so that items sharing a parent in the hierarchy are close to each other (embedding the ``horizontal'' structure), we obtain Poincar\'e embeddings \cite{nickel2017poincare}. More formally, this means that embedding norm represents depth in the hierarchy, and distance between embeddings the similarity of the respective items.

Figure \ref{fig:wnet} shows the results of embedding the WordNet noun hierarchy in two-dimensional Euclidean space (left) and the Poincar\'e disk (right). In the hyperbolic model, the types tend to be located near the boundary of the disk. In this region the space grows exponentially, which allows related types to be placed near one another and far from unrelated ones. The actual distance in this model is not the one visualized in the figure but the one given by Equation \ref{eq:hyperDist}.


\section{Entity Typing in Hyperbolic Space}
\label{sec:ETinHyperbolicSpace}

\subsection{Task Definition}
The task we consider is, given a context sentence $c$ containing an entity mention $m$, predict the correct type labels $t_m$ that describe $m$ from a type inventory $T$, which includes more than 10,000 types \cite{choi2018ultra}. The mention $m$ can be a named entity, a nominal, or a pronoun. The ground-truth type set $t_m$ may contain multiple types, making the task a multi-label classification problem.

\subsection{Objective}
We aim to analyze the effects of hyperbolic and Euclidean spaces when modeling hierarchical information present in the type inventory, for the task of fine-grained entity typing. Since hyperbolic geometry is naturally equipped to model hierarchical structures, we hypothesize that this enhanced representation will result in superior performance.
With the goal of examining the relation between the metric space and the hierarchy, we propose a regression model. We learn a function that maps feature representations of a mention and its context onto a vector space such that the instances are embedded closer to their target types.

The ground-truth type set contains a varying number of types per instance. In our regression setup, however, we aim to predict a fixed amount of labels for all the instances. This imposes strong upper bounds to the performance of our proposed model. Nonetheless, as the strict accuracy of state-of-the-art methods for the Ultra-Fine dataset is below 40\% \cite{choi2018ultra, xiong2019inductiveBias}, the evaluation we perform is still informative in qualitative terms, and enables us to gain better intuitions with regard to embedding hierarchical structures in different metric spaces.

\subsection{Method}
\label{subsec:methodology}

Given the encoded feature representations of a mention $m$ and its context $c$, noted as $e(m,c) \in \mathbb{R}^{n'}$ our goal is to learn a mapping function $f: \mathbb{R}^{n'} \rightarrow \mathcal{S}^n$, where $\mathcal{S}^n$ is the target vector space. We intend to approximate embeddings of the type labels $t_m$, previously projected into the space.
Subsequently, we perform a search of the nearest type embeddings of the embedded representation in order to assign the categorical label corresponding to the mention within that context. Figure~\ref{fig:overview} presents an overview of the model. 

The label distribution on the dataset is diverse and fine-grained. Each instances is annotated with three levels of granularity, namely \textit{coarse}, \textit{fine} and \textit{ultra-fine}, and on the development and test set there are, on average, five labels per item. This poses a challenging problem for learning and predicting with only one projection. As a solution, we propose three different projection functions, $f_{coarse}, f_{fine}$, and $f_{ultra}$, each one of them fine-tuned to predict labels of a specific granularity.

We hypothesize that the complexity of the projection increases as the granularity becomes \textit{finer}, given that the target label space per granularity increases. Inspired by \newcite{sanh2019hierarchicalEmbeddding}, we arrange the three projections in a hierarchical manner that reflects these difficulties. The \textit{coarse} projection task is set at the bottom layer of the model and more complex (\textit{finer}) interactions at higher layers. With the projected embedding of each layer, we aim to introduce an inductive bias in the next projection that will help to guide it into the correct region of the space. Nevertheless, we use shortcut connections so that top layers can have access to the encoder layer representation.

\subsection{Mention and Context Representations}

To encode the context $c$ containing the mention $m$, we apply the encoder schema of \newcite{choi2018ultra} based on \newcite{shimaoka2016attentive}. 
We replace the location embedding of the original encoder with a word position embedding $p_i$ to reflect relative distances between the $i$-th word and the entity mention. This modification induces a bias on the attention layer to focus less on the mention and more on the context. Finally we apply a standard Bi-LSTM and a self-attentive encoder \cite{mccann2017selfattentive} on top to get the context representation $C \in \mathbb{R}^{d_c} $.

For the mention representation we derive features from a character-level CNN, concatenate them with the Glove word embeddings \cite{pennington2014glove} of the mention, and combine them with a similar self-attentive encoder. The mention representation is denoted as $M \in \mathbb{R}^{d_m}$. The final representation is achieved by the concatenation of mention and context $[M;C] \in \mathbb{R}^{d_m + d_c}$.

\subsection{Projecting into the Ball}
\label{subsec:projecting-into-the-ball}

To learn a projection function that embeds our feature representation in the target space, we apply a variation of the re-parameterization technique introduced in \newcite{dhingra2018embeddingTextInHS}. The re-parameterization involves computing a direction vector $r$ and a norm magnitude $\lambda$ from $e(m,c)$ as follows:
\vspace{-2mm}
\begin{equation}
\label{eq:reparam}
\begin{split}
\overline{r}=\varphi_{dir}(e(m,c)), \quad r = \frac{\overline{r}}{\|\overline{r}\|}, \\
\overline{\lambda}=\varphi_{norm}(e(m,c)), \quad \lambda = \sigma(\overline{\lambda}),
\end{split}
\end{equation}

where $\varphi_{dir} : \mathbb{R}^{n'} \rightarrow \mathbb{R}^{n}$, $\varphi_{norm} : \mathbb{R}^{n'} \rightarrow \mathbb{R}$ can be arbitrary functions, whose parameters will be optimized during training, and $\sigma$ is the sigmoid function that ensures the resulting norm $\lambda \in (0, 1)$. The re-parameterized embedding is defined as $v = \lambda r$, which lies in $\mathcal{S}^n$.

By making use of this simple technique, the embeddings are guaranteed to lie in the Poincar\'e ball. This avoids the need to correct the gradient or the utilization of Riemannian-SGD \cite{bonnabel2011rsgd}. Instead, it allows the use of any optimization method in deep learning, such as Adam \cite{kingma2014Adam}. 

We parameterize the direction function $\varphi_{dir}: \mathbb{R}^{d_m + d_c} \rightarrow \mathbb{R}^n$ as a multi-layer perceptron (MLP) with a single hidden layer, using rectified linear units (ReLU) as nonlinearity, and dropout. We do not apply the ReLU function after the output layer in order to allow negative values as components of the direction vector. For the norm magnitude function $\varphi_{norm}: \mathbb{R}^{d_m + d_c} \rightarrow \mathbb{R}$ we use a single linear layer.

\subsection{Optimization of the Model} 
\label{subsec:optimization}

We aim to find projection functions $f_{i}$ that embed the instance representations closer to the respective target types, in a given vector space $\mathcal{S}^n$. As target space $\mathcal{S}^n$ we use the Poincar\'e Ball $\mathcal{B}^n$ and compare it with the Euclidean unit ball $\mathbb{R}^n$. Both $\mathcal{B}^n$ and $\mathbb{R}^n$ are metric spaces, therefore they are equipped with a distance function, namely the hyperbolic distance $d_{H}$ defined in Equation~\ref{eq:hyperDist}, and the Euclidean distance $d_{E}$ respectively, which we intend to minimize. Moreover, since the Poincar\'e Model is a conformal model of the hyperbolic space, \textit{i.e.} the angles between Euclidean and hyperbolic vectors are equal, the cosine distance $d_{\cos}$ can be used, as well.


We propose to minimize a combination of the distance defined by each metric space and the cosine distance to approximate the embeddings. Although formally this is not a distance metric since it does not satisfy the Cauchy-Schwarz inequality, it provides a very strong signal to approximate the target embeddings accounting for the main concepts modeled in the representation: \textit{relatedness}, captured via the distance and orientation in the space, and \textit{generality}, via the norm of the embeddings. 

To mitigate the instability in the derivative of the hyperbolic distance\footnote{$\lim_{y \to x} \partial_{x} |d_{H}(x,y)| \to \infty \ \forall x \in \mathcal{B}^n $} we follow the approach proposed in \newcite{deSa18tradeoffs} and minimize the square of the distance, which does have a continuous derivative in $\mathcal{B}^n$. Thus, in the Poincar\'e Model we minimize the distance for two points $u,v \in \mathcal{B}^n$ defined as:
\begin{equation}
\label{eq:poindist}
d_{\mathcal{B}}(u,v) = \alpha (d_{H}(u,v))^2 + \beta d_{\cos}(u,v)
\end{equation}

Whereas in the Euclidean space, for $x,y \in \mathbb{R}^n$ we minimize:
\begin{equation}
\label{eq:eudist}
d_{\mathbb{R}}(x,y) = \alpha d_{E}(x,y) + \beta d_{\cos}(x,y)
\end{equation}

The hyperparameters $\alpha$ and $\beta$ are added to compensate the bounded image of the cosine distance function in $[0,1]$.

\section{Hierarchical Type Inventories}
\label{sec:hierarchies}

In this section, we investigate two methods for deriving a hierarchical structure for a given type inventory. First, we introduce the datasets on which we perform our study since we exploit some of their characteristics to construct a hierarchy.

\subsection{Data}

\begin{table}[!t]
\begin{center}
\small
\begin{tabular}{lrrr}
\specialrule{.1em}{.05em}{.05em}
\textbf{Split} & \textbf{Coarse}  & \textbf{Fine}    & \textbf{Ultra-fine} \\
\hline
Train   & 2,416,593 & 4,146,143 & 3,997,318 \\
Dev     & 1,918    & 1,289    & 7,594 \\
Test    & 1,904    & 1,318    & 7,511 \\
\specialrule{.1em}{.05em}{.05em}
\end{tabular}
\caption{Type instances in the dataset grouped by split and granularity.}
\label{tab:type_count}
\end{center}
\vspace{-4mm}
\end{table}

We focus our analysis on the the Ultra-Fine entity typing dataset introduced in \newcite{choi2018ultra}. Its design goals were to increase the diversity and coverage entity type annotations. It contains 10,331 target types defined as free-form noun phrases and divided in three levels of granularity: \textit{coarse}, \textit{fine} and \textit{ultra-fine}. The data consist of 6,000 crowdsourced examples and approximately 6M training samples in the open-source version\footnote{\newcite{choi2018ultra} uses the licensed Gigaword to build part of the dataset resulting in about 25.2M training samples.}, automatically extracted with distant supervision, by entity linking and nominal head word extraction. Our evaluation is done on the original crowdsourced dev/test splits.

To gain a better understanding of the proposed model under different geometries, we also experiment on the OntoNotes dataset \cite{gillick2014context} as it is a standard benchmark for entity typing.

\subsection{Deriving the Hierarchies}
The two methods we analyze to derive a hierarchical structure from the type inventory are the following.

\noindent \textbf{Knowledge base alignment:} Hierarchical information can be provided explicitly, by aligning the type labels to a knowledge base schema. In this case the types follow the tree-like structure of the ontology curated by experts. On the Ultra-Fine dataset, the type vocabulary $T$ (\textit{i.e.} noun phrases) is extracted from WordNet \cite{miller1995wordnet}. Nouns in WordNet are organized into a deep hierarchy, defined by hypernym or ``IS A'' relationships. By aligning the type labels to the hypernym structure existing in WordNet, we obtain a type hierarchy. In this case, all paths lead to the root type \textit{entity}. In the OntoNotes dataset the annotations follow a pre-established, much smaller, hierarchical taxonomy based on ``IS A'' relations, as well.

\noindent \textbf{Type co-occurrences:} Although in practical scenarios hierarchical information may not always be available, the distribution of types has an implicit hierarchy that can be inferred automatically. If we model the ground-truth labels as nodes of a graph, its adjacency matrix can be drawn and weighted by considering the co-occurrences on each instance. That is, if $t_1$ and $t_2$ are annotated as true types for a training instance, we add an edge between both types. To weigh the edge we explore two variants: the frequency of observed instances where this co-relation holds, and the \textit{pointwise mutual information} ($pmi$), as a measure of the association between the two types\footnote{We adapt $pmi$ in order to satisfy the condition of non-negativity.}. 
By mining type co-occurrences present in the dataset as an affinity score, the hierarchy can be inferred. This method alleviates the need for a type inventory explicitly aligned to an ontology or pre-defined label correlations.

To embed the target type representations into the different metric spaces we make use of the library \textit{Hype}\footnote{\url{https://github.com/facebookresearch/poincare-embeddings/}} \cite{nickel2018lorentz}. This library allows us to embed graphs into low-dimensional continuous spaces with different metrics, such as hyperbolic or Euclidean, ensuring that related objects are closer to each other in the space. The learned embeddings capture notions of both similarity, through the relative distance among each other, and hierarchy, through the distance to the origin, i.e.\ the norm. The projection of the hierarchy derived from WordNet is depicted in Figure~\ref{fig:wnet}.



\section{Experiments}
\label{sec:experiments}

\begin{table*}[hbt!]
\small
\centering
\adjustbox{max width=\textwidth}{
\begin{tabular}{llrrrrrr|rrrr}
\specialrule{.1em}{.05em}{.05em}
\multirow{2}{*}{\textbf{Model}} & \multirow{2}{*}{\textbf{Space}} & \multicolumn{2}{c}{\textbf{Coarse}} & \multicolumn{2}{c}{\textbf{Fine}} & \multicolumn{2}{c}{\textbf{Ultra-fine}} & \multicolumn{2}{c}{\textbf{\begin{tabular}[c]{@{}c@{}}Coarse \\ + Ultra\end{tabular}}} & \multicolumn{2}{c}{\textbf{Variation}} \\
 &  & \textbf{MaF1} & \textbf{MiF1} & \textbf{MaF1} & \textbf{MiF1} & \textbf{MaF1} & \textbf{MiF1} & \textbf{MaF1} & \textbf{MiF1} & \textbf{MaF1} & \textbf{MiF1} \\
 \hline
\textsc{MultiTask} & - & 60.6 & 58.0 & 37.8 & 34.7 & 13.6 & 11.7 & - & - & - & - \\
 \hline
 \multirow{2}{*}{\textsc{WordNet}} & Hyper & 45.9 & 44.3 & 22.5 & 21.5 & 7.0 & 6.7 & 41.8 & 37.2 & -4.1 & -7.1 \\
 & Euclid & 56.1 & 54.2 & 26.6 & 25.3 & 7.2 & 6.5 & 56.6 & 48.5 & 0.6 & -5.7 \\
 \hline
\multirow{2}{*}{\begin{tabular}[c]{@{}l@{}}\textsc{WordNet}\\  \textsc{+ freq}\end{tabular}} & Hyper & 54.6 & 52.8 & 18.4 & 18.0 & 11.3 & 10.8 & 46.5 & 40.6 & -8.0 & -12.2 \\
 & Euclid & \textbf{56.7} & \textbf{54.9} & \textbf{27.3} & \textbf{26.0} & 12.1 & 11.5 & 55.8 & 49.1 & -0.9 & -5.8 \\
 \hline
\multirow{2}{*}{\textsc{freq}} & Hyper & 56.5 & 54.6 & 26.8 & 25.7 & \textbf{16.0} & 15.2 & 59.7 & \textbf{53.5} & 3.2 & \textbf{-1.1} \\
 & Euclid & 56.1 & 54.2 & 25.8 & 24.4 & 12.1 & 11.4 & \textbf{60.0} & 53.0 & \textbf{3.9} & -1.3 \\
 \hline
\multirow{2}{*}{\textsc{pmi}} & Hyper & 54.7 & 53.0 & 26.9 & 25.8 & \textbf{16.0} & \textbf{15.4} & 57.5 & 51.8 & 2.8 & -1.2 \\
 & Euclid & 56.5 & 54.6 & 26.9 & 25.6 & 12.2 & 11.5 & 59.7 & 53.0 & 3.2 & -1.5 \\
\specialrule{.1em}{.05em}{.05em}
\end{tabular}
}

\vspace{-2mm}
\subfloat[\label{tab:gran-res} Results on the same three granularities analyzed by \newcite{choi2018ultra}.]{\hspace{.6\linewidth}}
\hspace{.05\textwidth}
\subfloat[\label{tab:co-uf-res} Comparison to previous \textit{coarse} results.]{\hspace{.3\linewidth}}

\caption{Results on the test set for different hierarchies and spaces. The best results of our models are marked in bold. On (b) we report the comparison of adding the closest \textit{coarse} label to the \textit{ultra-fine} prediction, with respect to the \textit{coarse} results on (a).}
\label{tab:all-results}
\end{table*}

We perform experiments on the Ultra-Fine \cite{choi2018ultra} and OntoNotes \cite{gillick2014context} datasets to evaluate which kind of hierarchical information is better suited for entity typing, and under which geometry the hierarchy can be better exploited.

\subsection{Setup}
For evaluation we run experiments on the Ultra-Fine dataset with our model projecting onto the hyperbolic space, and compare to the same setting in Euclidean space. The type embeddings are created based on the following hierarchical structures derived from the dataset: the type vocabulary aligned to the WordNet hierarchy (\textsc{WordNet}), type co-occurrence frequency (\textsc{freq}), \textit{pointwise mutual information} among types (\textsc{pmi}), and finally, the combination of WordNet's transitive closure of each type with the co-occurrence frequency graph (\textsc{WordNet + freq}).
  
We compare our model to the multi-task model of \newcite{choi2018ultra} trained on the open-source version of their dataset (\textsc{MultiTask}).
The final type predictions consist of the closest neighbor from the \textit{coarse} and \textit{fine} projections, and the three closest neighbors from the \textit{ultra-fine} projection. 
We report Loose Macro-averaged and Loose Micro-averaged F1 metrics computed from the precision/recall scores over the same three granularities established by \newcite{choi2018ultra}.
For all models we optimize Macro-averaged F1 on \textit{coarse} types on the validation set, and evaluate on the test set. All experiments project onto a target space of 10 dimensions. The complete set of hyperparameters is detailed in the Appendix.

\section{Results and Discussion}

\subsection{Comparison of the Hierarchies}

\begin{table*}[hbt!]
\centering
\small
\adjustbox{max width=\textwidth}{
\begin{tabular}{lp{0.74\textwidth}}
\specialrule{.1em}{.05em}{.05em}
a) Example & Rin, Kohaku and Sesshomaru Rin befriends Kohaku, the demonslayer Sango's younger brother, while Kohaku acts as her guard when Naraku is using her for bait to lure Sesshomaru into \textbf{battle}. \\
Annotation & \textit{event, conflict, war, fight, battle, struggle, dispute, group\_action} \\
Prediction & \textsc{freq}: \textcolor{blue}{event, conflict, war, fight, battle}; \newline \textsc{WordNet}: \textcolor{blue}{event, conflict,} difference, engagement, assault \\
  \hline
b) Example & \textbf{The UN mission in Afghanistan} dispatched its own investigation, expressing concern about reports of civilian casualties and calling for them to be properly cared for. \\
Annotation & \textit{organization, team, mission} \\
Prediction & \textsc{freq}: \textcolor{blue}{organization, team, mission}, activity, operation; \newline \textsc{WordNet}: group, institution, branch, delegation, corporation \\
\hline
c) Example & Brazilian President Luiz Inacio Lula da Silva and Turkish Prime Minister Recep Tayyip Erdogan talked about designing a strategy different from sanctions at \textbf{a meeting} Monday, Amorim said. \\
Annotation & \textit{event, meeting, conference, gathering, summit} \\
Prediction & \textsc{freq}: \textcolor{blue}{event, meeting, conference}, film, campaign; \newline \textsc{WordNet}: entity, \textcolor{blue}{meeting, gathering}, structure, court \\
 \specialrule{.1em}{.05em}{.05em}
\end{tabular}
\caption{Qualitative analysis of instances taken from the development set. The predictions are generated with the hyperbolic models of \textsc{freq} and \textsc{WordNet}. Correct predictions are marked in blue color.}
\label{tab:qual-analysis}
}
\vspace{-2mm}
\end{table*}

Results on the test set are reported in Table~\ref{tab:all-results}. From comparing the different strategies to derive the hierarchies, we can see that \textsc{freq} and \textsc{pmi} substantially outperform \textsc{MultiTask} on the \textit{ultra-fine} granularity ($17.5\%$ and $29.8\%$ relative improvement in Macro F1 and Micro F1, respectively, with the hyperbolic model). 
Both hierarchies show a substantially better performance over 
the \textsc{WordNet} hierarchy on this granularity as well (MaF1 $16.0$ and MiF1 $15.4$ for \textsc{pmi} vs $7.0$ and $6.7$ for \textsc{WordNet} on the Hyperbolic model), indicating that these structures, created solely from the dataset statistics, better reflect the type distribution in the annotations. On \textsc{freq} and \textsc{pmi}, types that frequently co-occur on the training set are located closer to each other, improving the prediction based on nearest neighbor.

All the hierarchies show very low performance on \textit{fine} when compared to the \textsc{MultiTask} model. This exhibits a weakness of our regression setup. On the test set there are 1,998 instances but only 1,318 \textit{fine} labels as ground truth (see Table~\ref{tab:type_count}). 
By forcing a prediction on the \textit{fine} level for all instances, precision decreases notably. More details in Section~\ref{sub:regression}.


The combined hierarchy \textsc{WordNet + freq} achieves marginal improvements on \textit{coarse} and \textit{fine} granularities, while it degrades the performance on \textit{ultra-fine} when compared to \textsc{freq}.

By imposing a hierarchical structure over the type vocabulary we can infer types that are located higher up in the hierarchy from the predictions of the lower ones. To analyze this, we add the closest \textit{coarse} label to the \textit{ultra-fine} prediction of each instance. Results are reported in Table~\ref{tab:co-uf-res}. The improvements are noticeable on the Macro score (up to $3.9$ F1 points difference on \textsc{freq}) whereas Micro decreases. Since we are adding types to the prediction, this technique improves recall and penalizes precision. Macro is computed on the entity level, while Micro provides an overall score, showing that per instance the prediction tends to be better. The improvements can be observed on \textsc{freq} and \textsc{pmi} given that their predictions over \textit{ultra-fine} types are better.

\subsection{Comparison of the Spaces}

\begin{figure}[b]
\centering
	\includegraphics[width=0.9\linewidth,keepaspectratio]{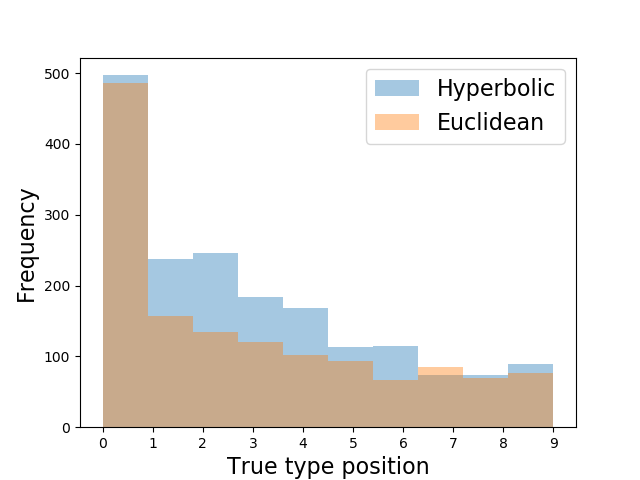}
	\caption{Histogram of ground-truth type neighbor positions for \textit{ultra-fine} predictions in Hyperbolic and Euclidean spaces on the test set.}
	\label{fig:neigh-cmp}
\end{figure}

When comparing performances with respect to the metric spaces, the hyperbolic models for \textsc{pmi} and \textsc{freq} outperform all other models on \textit{ultra-fine} granularity. Compared to its Euclidean counterpart, \textsc{pmi} brings considerable improvements ($16.0$ vs $12.2$ and $15.4$ vs $11.5$ for Macro and Micro F1 respectively). This can be explained by the exponential growth of this space towards the boundary of the ball, combined with a representation that reflects the type co-occurrences in the dataset. Figure~\ref{fig:neigh-cmp} shows a histogram of the distribution of ground-truth types as closest neighbors to the prediction.

On both Euclidean and hyperbolic models, the type embeddings for \textit{coarse} and \textit{fine} labels are located closer to the origin of the space. In this region, the spaces show a much more similar behavior in terms of the distance calculation, and this similarity is reflected on the results as well. 

The low performance of the hyperbolic model of \textsc{WordNet} on \textit{coarse} can be explained by the fact that \textit{entity} is the root node of the hierarchy, therefore it is located closer to the center of the space. Elements placed in the vicinity of the origin have a norm closer to zero, thus their distance to other types tends to be shorter (does not grow exponentially). This often misleads the model into assign \textit{entity} as the \textit{coarse}. See Table~\ref{tab:qual-analysis}c for an example.

This issue is alleviated on \textsc{WordNet + freq}. Nevertheless, it appears again when using the \textit{ultra-fine} prediction to infer the \textit{coarse} label. The drop in performance can be seen in Table~\ref{tab:co-uf-res}: Macro F1 decreases by $8.0$ and Micro F1 by $12.2$.

\subsection{Error analysis}
\label{sub:regression}
We perform an error analysis on samples from the development set and predictions from two of our proposed hyperbolic models. We show three examples in Table~\ref{tab:qual-analysis}. Overall we can see that predictions are reasonable, suggesting synonyms or related words. 

In the proposed regression setup, we predict a fixed amount of labels per instance. This schema has drawbacks as shown in example a), where all predicted types by the \textsc{freq} model are correct though we can not predict more, and b), where we predict more related types that are not part of the annotations.

In examples b) and c) we see how the \textsc{freq} model predicts the \textit{coarse} type correctly whereas the model that uses the WordNet hierarchy predicts \textit{group} and \textit{entity} since these labels are considered more general (\textit{organization} IS A \textit{group}) 
thus located closer to the origin of the space. 

To analyse precision and recall more accurately, we compare our model to the one of  \newcite{shimaoka2016attentive} (\textsc{AttNER}) and the multi-task model of \newcite{choi2018ultra} (\textsc{multi}). We show the results for macro-averaged metrics in Table~\ref{tab:setup-analysis}. Our model is able to achieve higher recall but lower precision. Nonetheless we are able to outperform \textsc{AttNER} with a regression model even though they apply a classifier to the task.

\begin{table}[!t]
\centering
\small
\begin{tabular}{p{1.2cm}p{0.54cm}p{0.54cm}p{0.54cm}p{0.54cm}p{0.54cm}p{0.54cm}p{0.54cm}}
\specialrule{.1em}{.05em}{.05em}
\multirow{2}{*}{\textbf{Model}} & \multicolumn{3}{c}{\textbf{Dev}} & \multicolumn{3}{c}{\textbf{Test}} \\
 & \textbf{P} & \textbf{R} & \textbf{F1} & \textbf{P} & \textbf{R} & \textbf{F1} \\
 \hline
\textsc{AttNER} & \textbf{53.7} & 15.0 & 23.5 & \textbf{54.2} & 15.2 & 23.7 \\
\textsc{freq} & 24.8 & \textbf{25.9} & 25.4 & 25.6 & \textbf{26.8} & 26.2 \\
\textsc{multi} & 48.1 & 23.2 & \textbf{31.3} & 47.1 & 24.2 & \textbf{32.0} \\
\specialrule{.1em}{.05em}{.05em}
\end{tabular}
\caption{Combined performance over the three granularities. Results are extracted from \newcite{choi2018ultra}.}
\label{tab:setup-analysis}
\end{table}

\subsection{Analysis Case: OntoNotes}

\begin{table}[!t]
\centering
\small
\begin{tabular}{p{0.6cm}p{0.3cm}p{0.54cm}p{0.54cm}p{0.54cm}p{0.54cm}p{0.45cm}p{0.4cm}}
\specialrule{.1em}{.05em}{.05em}
\multicolumn{1}{c}{\multirow{2}{*}{\textbf{Model}}} & \multicolumn{1}{c}{\multirow{2}{*}{\textbf{Sp}}} & \multicolumn{2}{c}{\textbf{Coarse}} & \multicolumn{2}{c}{\textbf{Fine}} & \multicolumn{2}{c}{\textbf{Ultra}} \\
\multicolumn{1}{c}{} & \multicolumn{1}{c}{} & \textbf{Ma} & \textbf{Mi} & \textbf{Ma} & \textbf{Mi} & \textbf{Ma} & \textbf{Mi} \\
\hline
\multirow{2}{*}{\textsc{Onto}} & Hy & \textbf{83.0} & 81.9 & 24.0 & 23.9 & 2.0 & 2.0 \\
 & Eu & 82.2 & \textbf{82.2} & 28.8 & 28.7 & 2.4 & 2.4 \\
 \hline
\multirow{2}{*}{\textsc{Freq}} & Hy & 81.7 & 81.8 & 27.1 & 27.1 & \textbf{4.2} & \textbf{4.2} \\
 & Eu & 81.7 & 81.7 & \textbf{30.6} & \textbf{30.6} & 3.8 & 3.8 \\
 \specialrule{.1em}{.05em}{.05em}
\end{tabular}
\caption{Macro and micro F1 results on OntoNotes.}
\label{tab:onto-res}
\end{table}

To better understand the effects of the hierarchy and the metric spaces we also perform an evaluation on OntoNotes \cite{gillick2014context}. We compare the original hierarchy of the dataset (\textsc{Onto}), and one derived from the type co-occurrence frequency extracted from the data augmented by \newcite{choi2018ultra} with this type inventory. The results for the three granularities are presented in Table~\ref{tab:onto-res}.

The \textsc{freq} model on the hyperbolic geometry achieves the best performance for the \textit{ultra-fine} granularity, in accordance with the results on the Ultra-Fine dataset. In this case the improvements of the frequency-based hierarchy are not so remarkable when compared to the \textsc{onto} model given that the type inventory is much smaller, and the annotations follow a hierarchy where there is only one possible path for every label to its \textit{coarse} type.

The low results on the \textit{ultra-fine} granularity are due to the reduced multiplicity of the annotated types (See Table~\ref{tab:onto_stats}). Most instances have only one or two types, setting very restrictive upper bounds for this setup.

\section{Related Work}
\label{sec:relwork}

Type inventories for the task of fine-grained entity typing \cite{ling2012fine, gillick2014context, yosef2012hyena} have grown in size and complexity \cite{delCorro2015finet, murty2017typenet, choi2018ultra}.
Systems have tried to incorporate hierarchical information on the type distribution in different manners. 
\newcite{shimaoka2017neural} encode the hierarchy through a sparse matrix. 
\newcite{xuBarbosa2018hierarchyAware} model the relations through a hierarchy-aware loss function.
\newcite{ma2016labelEmbedding} and \newcite{abhishek2017jointLearning} learn embeddings for labels and feature representations into a joint space in order to facilitate information sharing among them. Our work resembles \newcite{xiong2019inductiveBias} since they derive hierarchical information in an unrestricted fashion, through type co-occurrence statistics from the dataset.
These models operate under Euclidean assumptions. Instead, we impose a hyperbolic geometry to enrich the hierarchical information. 


Hyperbolic spaces have been applied mostly on complex and social networks modeling \cite{krioukov2010hypernetworks, verbeek2016socialNetworks}. In the field of Natural Language Processing, they have been employed to learn embeddings for Question Answering \cite{tay2018hyperQA}, in Neural Machine Translation \cite{gulcehre2018hyperAttentionNet}, and to model language \cite{leimeister2018skipGramHyper, tifrea2018poincareGlove}. 
We build upon the work of \newcite{nickel2017poincare} on modeling hierarchical link structure of symbolic data and adapt it with the parameterization method proposed by \newcite{dhingra2018embeddingTextInHS} to cope with feature representations of text.

\section{Conclusions}
\label{sec:concl}


Incorporation of hierarchical information from large type inventories into neural models has become critical to improve performance. 
In this work we analyze expert-generated and data-driven hierarchies, and the geometrical properties provided by the choice of the vector space, in order to model this information.
Experiments on two different datasets show consistent improvements of hyperbolic embedding over Euclidean baselines on very fine-grained labels when the hierarchy reflects the annotated type distribution.


\section*{Acknowledgments}
We would like to thank the anonymous reviewers for their valuable comments and suggestions, and we also thank Ana Marasovi\'c, Mareike Pfeil, Todor Mihaylov and Mark-Christoph M\"uller for their helpful discussions.
This work has been supported by the German Research Foundation (DFG) as part of the Research Training Group Adaptive Preparation of Information from Heterogeneous Sources (AIPHES) under grant No. GRK 1994/1 and the Klaus Tschira Foundation, Heidelberg, Germany.

\bibliography{acl2019}
\bibliographystyle{acl_natbib}


\newpage
\appendix

\section{Appendix}
\label{sec:appendix}


\subsection{Hyperparameters}

Both hyperbolic and Euclidean models were trained with the following hyperparameters.

\begin{table}[!htbp]
\begin{tabular}{lr}
\specialrule{.1em}{.05em}{.05em}
\textbf{Parameter} & \textbf{Value} \\
\hline
Word embedding dim & 300 \\
Max mention tokens & 5 \\
Max mention chars & 25 \\
Context length (per side) & 10 \\
Char embedding dim & 50 \\
Position embedding dim & 25 \\
Context LSTM dim & 200 \\
Attention dim & 100 \\
Mention dropout & 0.5 \\
Context dropout & 0.2 \\
Max gradient norm & 10 \\
Projection hidden dim & 500 \\
Optimizer & Adam \\
Learning rate & 0.001 \\
Batch size & 1024 \\
Epochs & 50 \\
\specialrule{.1em}{.05em}{.05em}
\end{tabular}
\caption{Model hyperparameters.}
\end{table}

\subsection{Dataset statistics}

\begin{table}[!htbp]
\begin{center}
\small
\begin{tabular}{lrrrr}
\specialrule{.1em}{.05em}{.05em}
\textbf{Split} & \textbf{Samples} & \textbf{Coarse}  & \textbf{Fine}    & \textbf{Ultra-fine} \\
      \hline
Train & 6,240,105 & 2,148,669 & 2,664,933 & 3,368,607  \\
Dev        & 1,998    & 1,612    & 947     & 1,860 \\
Test      & 1,998     & 1,598    & 964     & 1,864 \\
\specialrule{.1em}{.05em}{.05em}
\end{tabular}
\caption{Amount of samples with at least one label of the granularity organized by split on Ultra-Fine Dataset.}
\label{tab:instance_count}
\end{center}
\end{table}

\begin{table}[!htbp]
\begin{center}
\small
\begin{tabular}{lllll}
\specialrule{.1em}{.05em}{.05em}
\multicolumn{1}{c}{\textbf{Split}} & \multicolumn{1}{c}{\textbf{Samples}} & \multicolumn{1}{c}{\textbf{Coarse}} & \multicolumn{1}{c}{\textbf{Fine}} & \multicolumn{1}{c}{\textbf{Ultra}} \\
\hline
Train & 793,487 & 828,840 & 735,162 & 301,006 \\
Dev & 2,202 & 2,337 & 869 & 76 \\
Test & 8,963 & 9,455 & 3,521 & 417 \\
\specialrule{.1em}{.05em}{.05em}
\caption{Samples and label distribution by split on OntoNotes Dataset.}
\label{tab:onto_stats}
\end{tabular}
\end{center}
\end{table}

\end{document}